\documentclass[acmtog, nonacm]{acmart}

\usepackage{wrapfig}

\AtBeginDocument{%
  }

\setcopyright{acmcopyright}
\copyrightyear{2023}
\acmYear{2023}
\acmDOI{XXXXXXX.XXXXXXX}

\acmJournal{TOG}
\acmVolume{37}
\acmNumber{4}
\acmArticle{111}
\acmMonth{8}

\begin{document}

\title{Real-Time Neural BRDF with Spherically Distributed Primitives}

\author{Yishun Dou}
\authornote{Both authors contributed equally to this research}
\affiliation{
  \institution{Huawei}
  \country{China}}
\email{yishun.dou@gmail.com}

\author{Zhong Zheng}
\authornotemark[1]
\affiliation{
  \institution{Huawei}
  \country{China}
}
\email{zhengzhong5@huawei.com}

\author{Qiaoqiao Jin}
\affiliation{
  \institution{Shanghai Jiao Tong University}
  \country{China}
}

\author{Bingbing Ni}
\authornote{Corresponding Author: Bingbing Ni}
\affiliation{
  \institution{Shanghai Jiao Tong University}
  \country{China}
}
\email{nibingbing@sjtu.com}

\author{Yugang Chen}
\affiliation{
  \institution{Huawei}
  \country{China}
}

\author{Junxiang Ke}
\affiliation{
  \institution{Huawei}
  \country{China}
}

\renewcommand{\shortauthors}{Dou et al.}

\begin{abstract}


We propose a novel compact and efficient neural BRDF offering highly versatile material representation, 
yet with very-light memory and neural computation consumption towards achieving real-time rendering.
The results in Figure~\ref{fig:teaser}, rendered at full HD resolution on a current desktop machine, 
show that our system achieves real-time rendering with a wide variety of appearances, 
which is approached by the following two designs. 
On the one hand, noting that bidirectional reflectance is distributed in 
a very sparse high-dimensional subspace, 
we propose to project the BRDF into two low-dimensional components, 
\emph{i.e.}, two hemisphere feature-grids for incoming and outgoing directions, respectively.
On the other hand, learnable neural reflectance primitives are distributed on 
our highly-tailored spherical surface grid, 
which offer informative features for each component and 
alleviate the conventional heavy feature learning network to a much smaller one, 
leading to very fast evaluation. 
These primitives are centrally stored in a codebook 
and can be shared across multiple grids and even across materials, 
based on the low-cost indices stored in material-specific spherical surface grids. 
Our neural BRDF, which is agnostic to the material,
provides a unified framework that can represent a variety of materials in consistent manner. 
Comprehensive experimental results on measured BRDF compression, 
Monte Carlo simulated BRDF acceleration, and extension to spatially varying effect 
demonstrate the superior quality and generalizability achieved by the proposed scheme.


\end{abstract}




\begin{teaserfigure}
  \includegraphics[width=\textwidth]{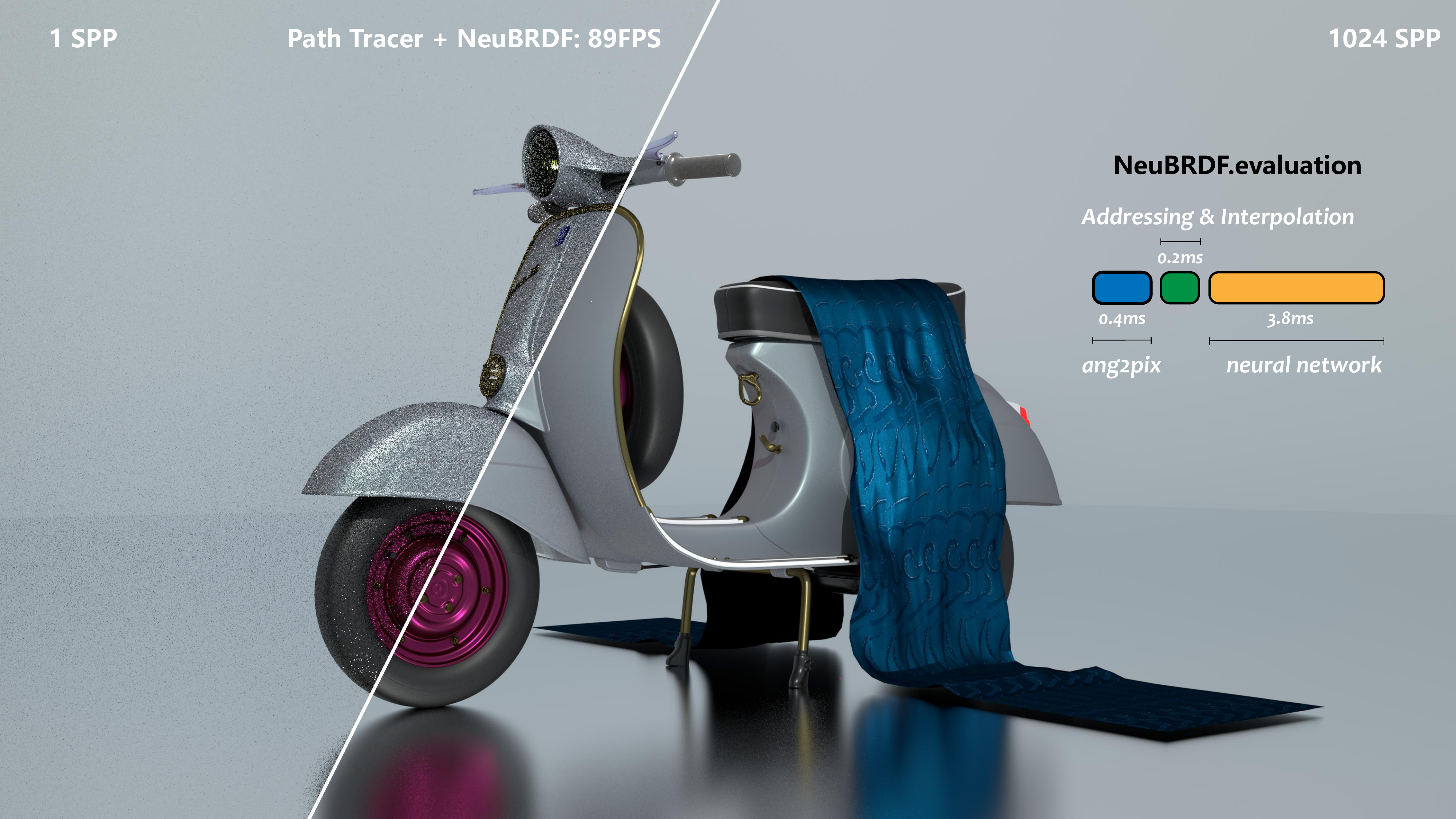}
  \caption{Our neural based reflectance model achieves real-time framerates 
  in current high-end desktop machine (i9 12900k and RTX 3090)
  and can cover a wide range of appearances. 
  We show the time breakdown of evaluating this reflectance model in a \textit{5-maxdepth} path tracer
  at full resolution (1 SPP, $1920 \times 1080$).}
  \label{fig:teaser}
\end{teaserfigure}

\maketitle

\section{Introduction}
\label{sec:intro}

Reflectance model is one of the most critical factors that affects the rendering photorealism.
A common choice for \textit{real-time} rendering is hand-crafted analytical 
BRDF models~\cite{walter2007microfacet, burley2012physically}, 
while they often struggle with reproducing realistic appearance. 
As for the methods offering high realism, measured BRDF models incur prohibitive storage and transmission overhead, 
especially for anisotropic and spatially varying materials; 
Monte Carlo simulation models rely on a large number of light paths, 
which prevents their popularity in real-time applications. 

One attractive choice is approximating the reflectance by implicit neural network~\cite{park2019deepsdf}, 
which has compact storage and can be potentially accelerated by the recent emerged 
feature-grid encoding~\cite{muller2022instant, takikawa2022variable}. 
However, the 6D input to the BRDF poses challenges for building such an acceleration structure. 
We propose to alleviate this difficulty through 
a \textit{factorization} which decomposes the BRDF to two 3D hemisphere feature-grids that are then decoded by a tiny MLP. 
Our system is designed according to the following principles: 
\begin{itemize}
  \item \textbf{Efficiency.} BRDF evaluation at shading point would executes million times on full HD resolution. 
  Operations on spherical feature-grid should be simple and the neural network should be kept lightweight. 
  \item \textbf{Mild memory cost.} The feature-grid is instance-specific and often trades memory for fast evaluation. 
  Hence, we should maximize the memory-quality tradeoff. 
  \item \textbf{Generalization.} We aim at providing a unified framework for a wide range of materials. 
  We strive for a solution that is agnostic to the materials of isotropic or anisotropic, measured or simulated, 
  spatially varying or not, and dense sampling or sparse.
\end{itemize}

The first principle spurs us to construct informative feature-grid that alleviates the heavy feature learning network 
to a much smaller one. 
It is also confronted with challenges in sphere pixelation. 
To this end, we exhaust commonly used sphere discretization methods, which are discussed and experimented in this paper. 
Moreover, the chosen HEALPix~\cite{gorski2005healpix} sphere pixelation is then tailored 
for constructing our reflectance feature-grid. 

The second principle poses challenges for the methods 
that directly store separate features at each grid~\cite{takikawa2021neural, muller2022instant}. 
To maximize the memory utilization, we propose to use neural reflectance primitives. 
These primitives are centrally stored in a codebook and can be shared across multiple grids and 
even across materials. 
As a result, the sphere grid is only required to store the low-cost index pointing to primitive 
and a material could be approximated by decoding a combination of these primitive codes,
\emph{i.e.} a good tradeoff between representation capability and memory compactness. 

Built on an auto-decoder framework, the factorization is implicitly achieved during the end-to-end joint training. 
Our \textbf{Neu}ral \textbf{BRDF} (NeuBRDF) enjoys the following properties: 
\begin{itemize}
  \item NeuBRDF represents measured or offline simulated materials effectively and 
  can be evaluated in \textit{real-time} at rendering. 
  \item NeuBRDF provides a \textit{unified framework} for a rich diversity of materials in a consistent manner. 
  \item Memory-quality tradeoffs can be easily achieved. 
  \item With a neural texture, NeuBRDF can be used for SVBRDF.
\end{itemize}

\medskip
\section{BACKGROUND AND RELATED WORK}

Our approach is related to many previous works in BRDF factorization, 
feature-based neural representations, neural reflectance models, and sphere pixelation. 
We explore the connection among the areas below.

\paragraph{BRDF Factorization}
Suitable analytic models are not always available for a desired effect, 
and directly tabulating reflectance data is prohibitive due to the high dimensionality. 
To overcome this limitation, 
some pioneers have proposed to factorize the BRDF function into lower dimensional factors, 
such as spherical harmonics~\cite{westin1992predicting}, 
Zernike polynomials~\cite{koenderink1996bidirectional} and so on. 
Factors are weighted summed or producted~\cite{mccool2001homomorphic, latta2002homomorphic} 
to approximate the high dimensional BRDF function. 
Similarly, Bagher et al.~\shortcite{bagher2016non} define a non-parametric factor microfacet model 
using tabluated definitions for three functional components (\textit{D}, \textit{G}, and \textit{F}). 
A commonality among these methods is they fall in linear transformation 
(or with one non-linearity). 
In contrast, our factorization achieved via auto-decoder learning is capable of modeling 
complex non-linear functions. 
In other words, the relationship between existing factorization and ours is similar to that 
of PCA and auto-encoder.

\paragraph{Feature-based Neural Representation}


Feature-based approaches use differentiable feature primitives to represent the scene 
including geometric shapes and materials. 
In terms of geometric shapes, feature-based approaches 
discretize the spatial space into a multiscale regular grid and store local features 
in an octree-based volume~\cite{takikawa2021neural}, a hash table~\cite{muller2022instant}, 
or a dictionary~\cite{takikawa2022variable}. 
Focusing on materials is much harder than geometric shapes since typically used 
BRDFs have a higher input dimension (6D) than the 3D geometry.
However, expanding the 3D spatial grid directly into higher dimensions to capture BRDFs 
is prohibitive for its large memory footprint.

\begin{figure*}[t]
  \centering
  \includegraphics[width=1.0\textwidth]{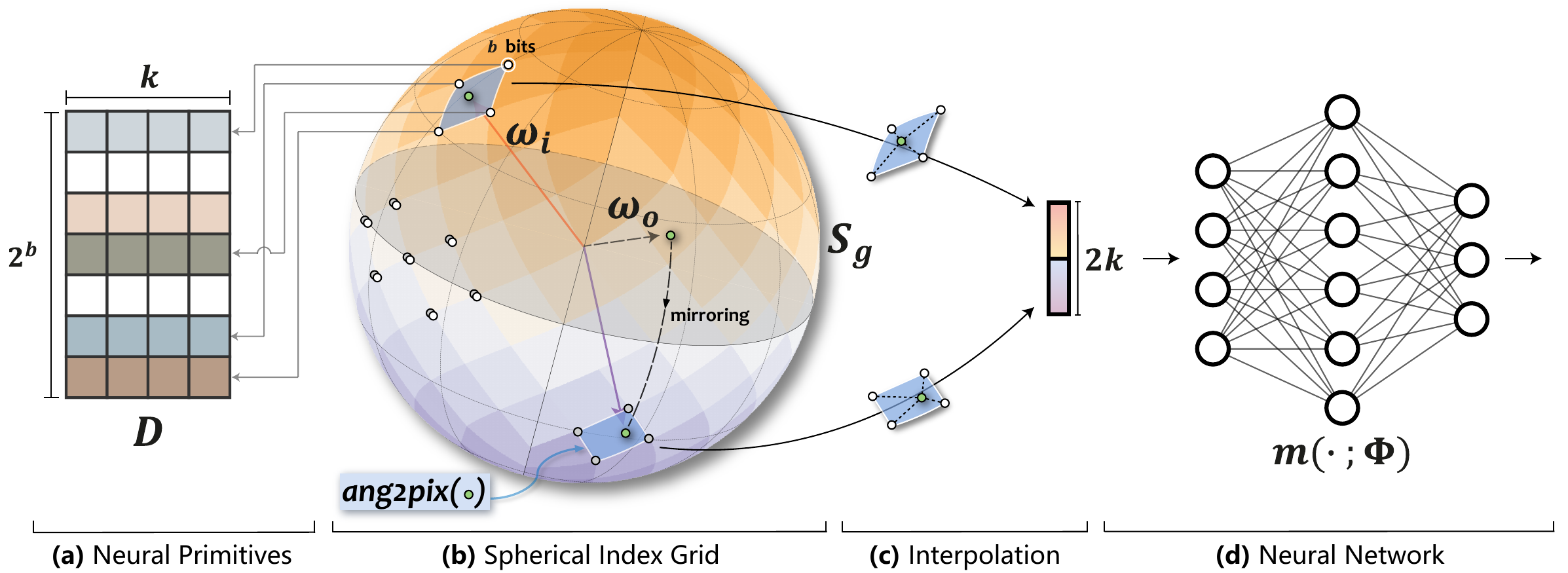}
  \caption{\textbf{Illustration of the neural BRDF with spherically distributed primitives.}
  \textbf{(a)} Learnable neural reflectance primitives are centrally stored on a codebook $\mathcal{D}$.
  \textbf{(b)} Given the incident direction $\boldsymbol{\omega}_i$ and outgoing $\boldsymbol{\omega}_o$, 
  we find the surrounding pixels independently on a spherical index grid $\mathcal{S}_g$. 
  For all resulting corner indices, 
  we look up the corresponding $k$-dimensional neural primitive from the codebook. 
  \textbf{(c)} Interpolating them according to the relative position of 
  $\boldsymbol{\omega}_i$ and  $\boldsymbol{\omega}_o$
  within the respective curvilinear quadrilateral pixels. 
  \textbf{(d)} A tiny MLP takes the concatenated features and outputs final bidirectional reflectance value.
  }
  \label{fig_pipeline}
\end{figure*}



\paragraph{Neural Reflectance Models}
There are two kinds of approaches for neural based SVBRDF / BTF / BRDF modeling. 
One kind of approaches use one neural network to represent one material. 
For example, \cite{sztrajman2021neural} used different lightweight networks 
to represent BRDFs of different materials, 
and the autoencoder architecture achieves compact representation. 
In order to represent and render a variety of material appearances at different scales, 
NeuMIP was proposed by~\cite{kuznetsov2021neumip}. 
\cite{rainer2019neural} used a latent vector to represent each texel 
as a neural representation of BTFs. 
Most recently, \cite{fan2023neural} approximated a BTF with a function of 2D 
spatial coordinates and half-vector coordinates. 
However, the half-vector poses a great risk of loss of bidirectional reflectance 
and thus requires a heavy MLP decoder for correction. 
The other kind of approaches use one neural network to represent a variety of materials. 
\cite{rainer2020unified} represented all materials with a unified model, 
which also uses the autoencoder architecture. 
But this approach may lead to a decline in the quality of representation. 
In order to realize the compact representation and keep the high quality of recovery, 
deepBRDF was proposed by~\cite{hu2020deepbrdf}, 
which uses a convolutional neural network. 
The neural process also was proposed by~\cite{zheng2021compact} 
for a compact representation of measured BRDFs. 
Recently, \cite{fan2022neural} proposed to perform layering in the 
neural space with the latent code compressed by a neural network. 


\paragraph{Sphere Pixelation}
Feature-grid relies on a well-characterised spatial pixelation, 
such as the octree for the neural SDF representation~\cite{takikawa2021neural}. 
Besides, representing materials spurs a demand of exploiting a sphere pixelation to 
accommodate spherical distributed features. 
We employ the Hierarchical Equal Area isoLatitude Pixelation 
(HEALPix)~\cite{gorski2005healpix}, 
which was designed for efficient and incremental discretization of full-sky maps 
in application to the satellite missions to measure the cosmic microwave background in astrophysics~\cite{perraudin2019deepsphere}. 
It provides a deterministic, uniform, and hierarchical sampling method for the sphere surface. 
Other sphere pixelation methods, such as 
longitude-latitude lattice, Layered Sukharev grid~\cite{sukharev1971optimal}, and 
Fibonacci lattice~\cite{gonzalez2010measurement}, 
are inadequate in term of efficiency or effectiveness, 
which will be discussed and experimented in this article.

\section{Method}
Our goal is to represent bidirectional reflectance distribution function (BRDF) using a neural framework
that maps bidirections to reflectance values and is evaluated in \textit{real-time} 
with \textit{mild memory cost}.  
The material is one of the most critical factors that affects the rendering photorealism, 
yet existing realistic approaches 
either incur large memory overhead or are computationally expensive.
We therefore opt for approximating it to keep the cost down. 


The key idea is to build a neural framework that consists of a
low-dimensional tailored feature-grid and a tiny neural network  
to approximate the high-dimensional bidirectional reflectance distributions.
In this section, we discuss the algorithmic choices 
that are key to satisfy the design principles outlined in Section~\ref{sec:intro}.
Implementation and practice consideration are discussed in Section~\ref{sec:implement}.


\subsection{Algorithm Overview}
Evaluating the NeuBRDF consists of \textbf{querying} a spherical index grid $\mathcal{S}_g$ 
by directions $\boldsymbol{\omega}_{i}$ and $\boldsymbol{\omega}_{o}$,  
\textbf{lookup} the codebook $\mathcal{D}$, 
and \textbf{inference} a tiny MLP; see Figure~\ref{fig_pipeline} for an illustration.

\paragraph{Fast Evaluation}
To enable effective and efficient BRDF evaluation at rendering,
we leverage the implicit field which bases on \textit{feature-grid}.
In contrast to the \textit{global} methods (\emph{e.g.} NeRF~\cite{mildenhall2021nerf})
that consist entirely of an MLP, the feature-grid allows the use of a much smaller network. 
However, it is prohibitive to expand the commonly used 3D spatial grid 
(\emph{e.g.} NGP~\cite{muller2022instant}) in scene representation 
directly into higher dimensions to capture BRDFs, 
since a 3D one already incurs a large memory footprint.


We consider that the bidirectional reflectance is distributed in a very sparse high-dimensional subspace and 
therefore propose to factorize the BRDF as a combination of low-dimensional components.
Concretely, we project the 6D BRDF onto two compact 3D hemispherical surface feature-grids 
along the outgoing $\boldsymbol{\omega}_{o}$ and incident $\boldsymbol{\omega}_{i}$ directions.
These two factors are used to compute the final reflectance via a tiny neural network, 
leading to efficient evaluation.

\paragraph{Memory Policy}
A tradeoff with feature-grid representations is that they can be quickly evaluated, 
but typically have a large memory overhead to cache the bulky features.
While our factorization-based solution offers dimensionality reduction, 
caching the 3D hemispherical features still incurs significant overhead.
Furthermore, the learned feature-grid is instance-specific,
which means that the memory cost grows with the number of materials in the rendering scenes.

To this end, we incorporate the vector quantization (VQ) compression technique.
The features that full of the hemisphere surface grid are replaced with indices into a learned codebook~\cite{takikawa2022variable}.
Notably, we could further maximize the memory-quality tradeoff
by reusing one codebook for an arbitrary number of materials,  
where each material has an exclusive indices grid.
We therefore term the prototype vectors in the codebook the \textit{neural reflectance primitives}.
These primitives, the indices, and the tiny MLP network are all trained jointly.


\begin{figure}[h]
  \centering
  \includegraphics[width=0.99\columnwidth]{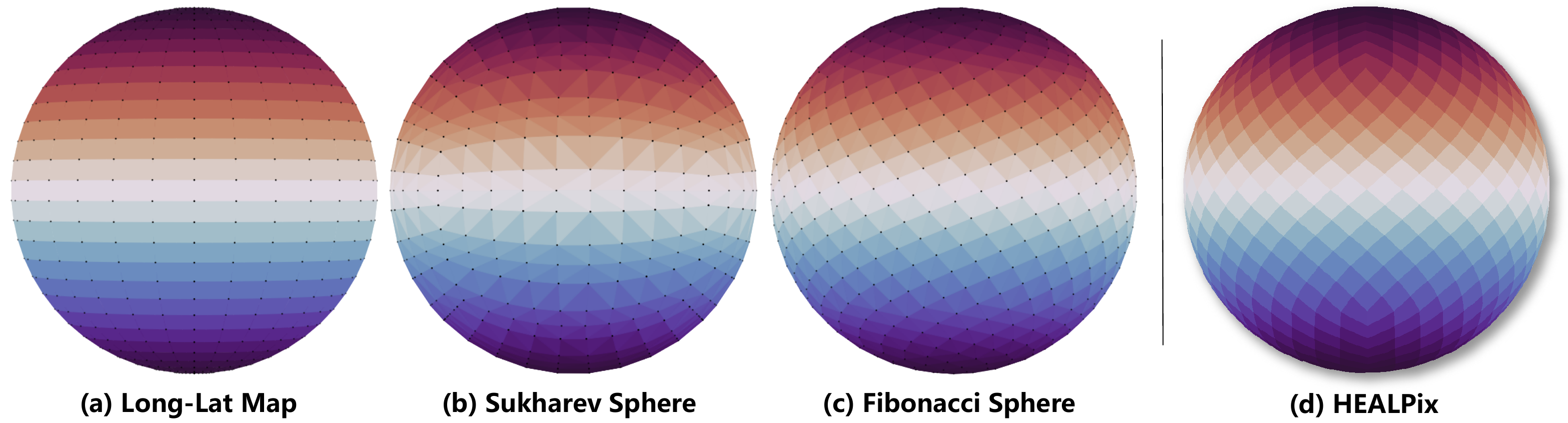}
  \caption{\textbf{Comparison among different sphere pixelations.} 
  }
  \label{fig_sphere_compare}
\end{figure}
\subsection{Spherical Feature-Grid}


Neural implicit fields in previous works are mainly used to represent low-dimensional signals, 
such as images and SDFs, thus typically a 3D grid is sufficient for the variants that base on feature-grid. 
Even for the application of representing 5D radiance fields,
the feature-grids are still built in 3D space, 
which represent the radiance by spherical harmonics~\cite{fridovich2022plenoxels} 
or only represent the Lambertian effect~\cite{muller2022instant}.  
In our case, the 6D BRDF is projected along
the outgoing $\boldsymbol{\omega}_o$ and incident $\boldsymbol{\omega}_i$ directions, 
resulting in two factors 
that are used to compute reflectance value.

Each factor is represented by features distributed on a hemisphere surface $\mathcal{H}$,
where the features are located in some discretized grid.
Although there are several spherical discretization methods, 
whether they meet the needs of real-time BRDF remains to be discussed.
Here we explore a structure that is appropriate for BRDF.


\subsubsection{Principles and Choices}
\label{sec:subsub_pixelation_principle}
Discretizing the sphere surface is much more complicated than that of the cube volume used in scene representation,
in terms of uniform pixelation and fast pixel indexing,
both of which are naturally achieved by the cube volume with regular discretization 
but pose challenges for the sphere surface.
To achieve the post pruning for sparse measurements, the property of hierarchy is required.
That is, unsupervised pixels should be replaced with their parent (macro) pixel, 
resulting in a multi-resolution sparse grid at runtime.
Together, the sphere discretization should meet the following main requirements:  
\begin{itemize}
  \vspace{-1mm}
  \item \textbf{Uniform Pixelation.} 
  Firstly, it conforms to the pixelation principle of image or voxel.
  Secondly, the sphere pixelation should be agnostic of materials and sampling strategies. 
  \item \textbf{Fast Indexing.} Real-time rendering requires fast indexing of sphere pixels
  in arbitrary direction.
  \item \textbf{Hierarchy.} For sparse measurements, a post grid pruning is necessary. 
  The intrinsic hierarchy is required to supported multi-resolution sparse grids.
  \vspace{-1mm}
\end{itemize}

There exist several sphere discretization choices, 
such as longitude-latitude lattice, Layered Sukharev grid~\cite{sukharev1971optimal}, 
Fibonacci lattice~\cite{gonzalez2010measurement}, 
and Hierarchical Equal Area isoLatitude Pixelation (HEALPix)~\cite{gorski2005healpix}.
Figure~\ref{fig_sphere_compare} illustrates the comparison of these choices. 
The simplest longitude-latitude lattice is suitable for indexing which is directly achieved by rounding up and down, 
but it arranges too many grid points on the polar cap.  
Sukharev grid also fails to achieve uniform pixelation. 
Although Fibonacci lattice can evenly distribute points on a sphere surface,  
it cannot locate the pixel for a given direction analytically and 
requires a nearest neighbor search,   
which prevents it from being used in real-time rendering. 
In contrast, HEALPix satisfies the above requirements; summarized in Table~\ref{tab:comp_pixelation}. 
We therefore employ HEALPix as the underlying sphere grid of our neural BRDF framework. 
Moreover, the intrinsic hierarchy of HEALPix facilitates grid pruning 
for isotropic reflectance or sparse measurements, which we will discuss in Section~\ref{sec:implement}.

\begin{table}  
  \caption{Comparison of pixelation methods. 
  \textbf{Indexing}: time consumption of
  obtaining the corresponding pixel $p$ given bidirection $\boldsymbol{\omega}_{i}$ and $\boldsymbol{\omega}_{o}$
  for a $1920 \times 1080$ frame. 
  \textbf{Iso-Lat}itude: an essential factor for consistent modeling of 
  the Fresnel term of reflectance at various azimuth angles.
  }
  \label{tab:comp_pixelation}
  \begin{minipage}{0.99\columnwidth}
  \begin{center}
  \begin{tabular}{l  c  c  c c}
    \toprule
    \textbf{Pixelation}  & \textbf{Uniform} & \textbf{Indexing} & \textbf{Hierarchy} & \textbf{Iso-Lat} \\ \hline
    \addlinespace[3pt]
    Long-Lat   & No        & $< 0.5$ms & No & Yes \\
    Sukharev   & No        & $\sim5$ms & No & No \\
    Fibonacci  & Yes       & $\sim5$ms & No & No \\
    HEALPix    & Yes       & $< 0.5$ms & Yes & Yes\\
    \bottomrule
  \end{tabular}
  \end{center}
  \centering
  \vspace{-2mm}
  \end{minipage}
  \vspace{-2mm}
\end{table}

\subsubsection{HEALPix-Based Data Structure}  
\label{sec:healpix_based_data_structure}

HEALPix tessellates the sphere into equal area curvilinear quadrilaterals.
The base resolution comprises 12 pixels in three rings around the poles and equator.
The resolution of the grid is determined only by one hyperparameter $N_{side}$,  
which defines the number of divisions along the side of a base-resolution pixel.
As a result, a HEALPix map has $12N_{side}^{2}$ pixels and $12N_{side}^{2} + 2$ grid points,
where the pixels have the same area. 

\begin{figure}[h]
  \centering
  \includegraphics[width=0.999\columnwidth]{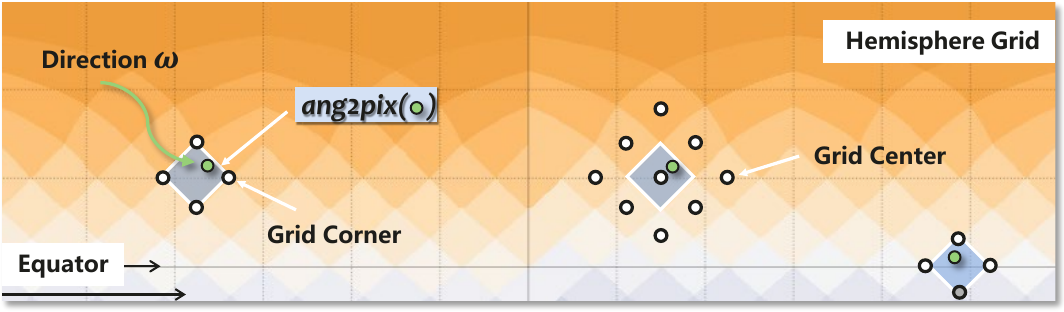}
  \caption{\textbf{Illustration of our hemisphere grid in Cartesian coordinate.} 
   Arranging features at grid corner ease the query of nearest neighboring features. 
   A hemisphere in our design includes pixels covering the equator.
  }
  \label{fig_healpix_orth}
  \vspace{-2mm}
\end{figure}

Unlike prior work that used HEALPix for cosmological applications~\cite{perraudin2019deepsphere},
we arrange features to points at grid corner instead of the pixel center;
Figure~\ref{fig_healpix_orth} shows the comparison.
This design reduces the computation cost of pixel indexing for four nearest neighboring features.
Although arranging features at the grid center has a modest indexing cost for nine nearest neighbors,
the number of memory accesses doubles due to the excessive number of features.

Consider a hemisphere covering equator line 
(\emph{i.e.} a ring adjacent to the equator line in another hemisphere is included).
The number of the hemisphere grid points is $n = 6N_{side}^2 + 6N_{side} + 1$
\footnote{A ring in equatorial belt comprises $4N_{side}$ points~\cite{gorski2005healpix}. 
The hemisphere excluding equator has $6N_{side}^2 - 2N_{side} + 1$ points.}. 
We denote a hemispherical feature-grid as $\mathcal{H}_g \in \mathbb{R}^{n \times k}$, 
where $k$ is the feature dimension.

Given a query direction $\boldsymbol{\omega} = (\theta, \phi)$, where $\theta \in [0, \pi]$ is the 
colatitude in radians measured from the north pole and $\phi$ is the longitude in radians measured eastward.
\textbf{(a)} We first find the curvilinear quadrilateral $p$ to which the direction $\boldsymbol{\omega}$ belongs,
through the fast analytic indexing $ang2pix$. We refer the readers to the reference~\cite{gorski2005healpix} for details.
\textbf{(b)} Then we extract the features $\boldsymbol{z}_{p}^{t} \in \mathcal{H}_g$ 
at four corners of the quadrilateral pixel $p$,
where $t \in \{ 1,2,3,4 \}$ is the local indices.
Meanwhile, we fetch the corresponding spherical coordinates $\boldsymbol{\omega}_{p}^{t}$ to compute the interpolation weights 
$w_p^{t}$ during training, where the weights are assigned with $1/4$ at runtime for efficiency:
\begin{equation}
  w_{p}^{t} = 
  \begin{cases}
  w_{p}^{t} / \sum_{j=1}^{4} w_{p}^{j}, \ \ \text{where} \ \ w_{p}^{t} = 1 / d_{p}^{t},  & \text{Training,} \\
  1 / 4, & \text{Runtime,}
  \end{cases}
\end{equation}
where $d_{p}^{t}$ is the great-circle distance between the spherical coordinates of 
$\boldsymbol{\omega}_{p}^{t}$ and $\boldsymbol{\omega}$.
The query for a hemispherical feature-grid $\mathcal{H}_g$ by $\boldsymbol{\omega}$ can be expressed as:
\begin{equation}
  \psi(\boldsymbol{\omega}; \mathcal{H}_g) = \sum_{t=1}^{4} w_{p}^{t} \boldsymbol{z}_{p}^{t},
  \ \ \text{where} \ \ p = ang2pix(\boldsymbol{\omega}; \mathcal{H}_g),
\end{equation}
\textbf{(c)} Recall that this process takes place independently for 
incident $\boldsymbol{\omega}_i$ and outgoing $\boldsymbol{\omega}_o$
and therefore two hemispherical feature-grids are required for representing these two factors.
Effectively, we merge the two hemispheres into a sphere that comprises three duplicate rings,
the equator and its two nearest adjacent rings. 
The number of the sphere grid points $2n$, \emph{i.e.} $12N_{side}^2 + 12N_{side} + 2$. 
We denote a spherical feature-grid as $\mathcal{S}_g \in \mathbb{R}^{2n \times k}$.
Eventually, querying a spherical grid $\mathcal{S}_g$ by $\boldsymbol{\omega}_{i}$ and $\boldsymbol{\omega}_{o}$ is:
\begin{align}
  \hat{\boldsymbol{\omega}}_{o} &= (\pi-\boldsymbol{\omega}_{o}.\theta, \boldsymbol{\omega}_{o}.\phi), \nonumber \\
  \psi(\boldsymbol{\omega}_{i}, \boldsymbol{\omega}_{o}; \mathcal{S}_g) &= 
  [\psi(\boldsymbol{\omega}_i; \mathcal{S}_g^{+}), \psi(\hat{\boldsymbol{\omega}}_o; \mathcal{S}_g^{-})],
\end{align}
where $[\cdot, \cdot]$ denotes concatenation. 
The $\mathcal{S}_g^{+}$ and $\mathcal{S}_g^{-}$ denote the north and south hemispherical feature-grid, 
both of which include the equator ring and a nearest adjacent ring to the equator in another hemisphere.

\subsubsection{Decoder}
In order to lift the partially low-dimensional factors to the original 6D space, 
we employ a multilayer perceptron (MLP) as a non-linear decoder,
where the input to MLP is the result of spherical feature-grid query 
$\psi(\boldsymbol{\omega}_{i}, \boldsymbol{\omega}_{o}; \mathcal{S}_g)$.
Then, the BRDF $f(\boldsymbol{\omega}_i, \boldsymbol{\omega}_o)$ of our method is described as:
\begin{equation}
  f(\boldsymbol{\omega}_i, \boldsymbol{\omega}_o) \approx 
  m(\psi(\boldsymbol{\omega}_{i}, \boldsymbol{\omega}_{o}; \mathcal{S}_g); \Phi),
\end{equation}
where $m(\cdot; \Phi)$ is the MLP with parameters $\Phi$ (including weights and biases).
Interestingly, the architecture designed from the perspective of factorizing the 6D BRDF
resembles that from the parametric encoding~\cite{peng2020convolutional, liu2020neural, takikawa2021neural, muller2022instant}.
The additional trainable parameters (beyond weights and biases)
allow the use of a tiny MLP without sacrificing representation quality and 
yield immediate \textit{benefits}: 
(1) NeuBRDF can be trained to convergence much faster than the method that consists entirely of MLPs; 
(2) It provides feasibility for applying neural BRDF to real-time rendering applications. 


\paragraph{Discussion}
The difference between parametric-encoding methods and ours is obvious in the intention of MLP.  
Conceptually, the MLP in our framework is used to \textit{compensate} the lossy factorization.
Whereas for methods with parametric encoding, 
the MLPs are typically used for local (interpolated) feature decoding~\cite{peng2020convolutional},
multi-resolution aggregation~\cite{takikawa2021neural}, 
or handling non-Lambertian effect~\cite{muller2022instant} for neural radiance application.



\subsection{Neural Reflectance Primitives}
\label{sec:neural_primitives}
While our factorization solution can perform dimensionality reduction, 
a spherical feature-grid $\mathcal{S}_g$ with our default setting still 
requires about $0.8$ million parameters ($N_{side}=64, k=16$). 
Following~\cite{takikawa2022variable}, 
we compress the storage by incorporating vector-quantization into our framework.
Concretely, the $\mathcal{S}_g \in \mathbb{R}^{2n \times k}$ is 
compressed into an integer vector $\mathcal{S}_g \in \mathbb{Z}^{2n}$ 
with the range $[0, 2^b - 1]$. 
The integers are used as \textit{indices} into a codebook matrix $\mathcal{D} \in \mathbb{R}^{2^{b} \times k}$,
where $b$ is the bitwidth to store such an integer. 
We thus now name $\mathcal{S}_g$ the spherical index grid.
The BRDF $f(\boldsymbol{\omega}_i, \boldsymbol{\omega}_o)$ is computed as: 
\begin{equation}
  f(\boldsymbol{\omega}_i, \boldsymbol{\omega}_o) \approx 
  m(\psi(\boldsymbol{\omega}_{i}, \boldsymbol{\omega}_{o}; \mathcal{D}[\mathcal{S}_g]); \Phi),
\end{equation}
where $[\cdot]$ is the indexing operation and $\mathcal{D}[\mathcal{S}_g]$ denotes lookup.
(The training of these indices is detailed in Section~\ref{sec:implement}.)
For \textit{fp16} storage,
this gives us a compression ratio of $16 \cdot 2n \cdot k / (2n \cdot b + 16 \cdot k \cdot 2^{b})$, 
which can be orders of magnitude when $b$ is small and $n$ is large. 

Notably, we provide an option to \textit{maximize} the quality-cost tradeoff
for deploying NeuBRDF into memory constrained devices.   
\begin{wrapfigure}{r}{0.32\columnwidth}
  \vspace{-4mm}
  \includegraphics[width=0.32\columnwidth]{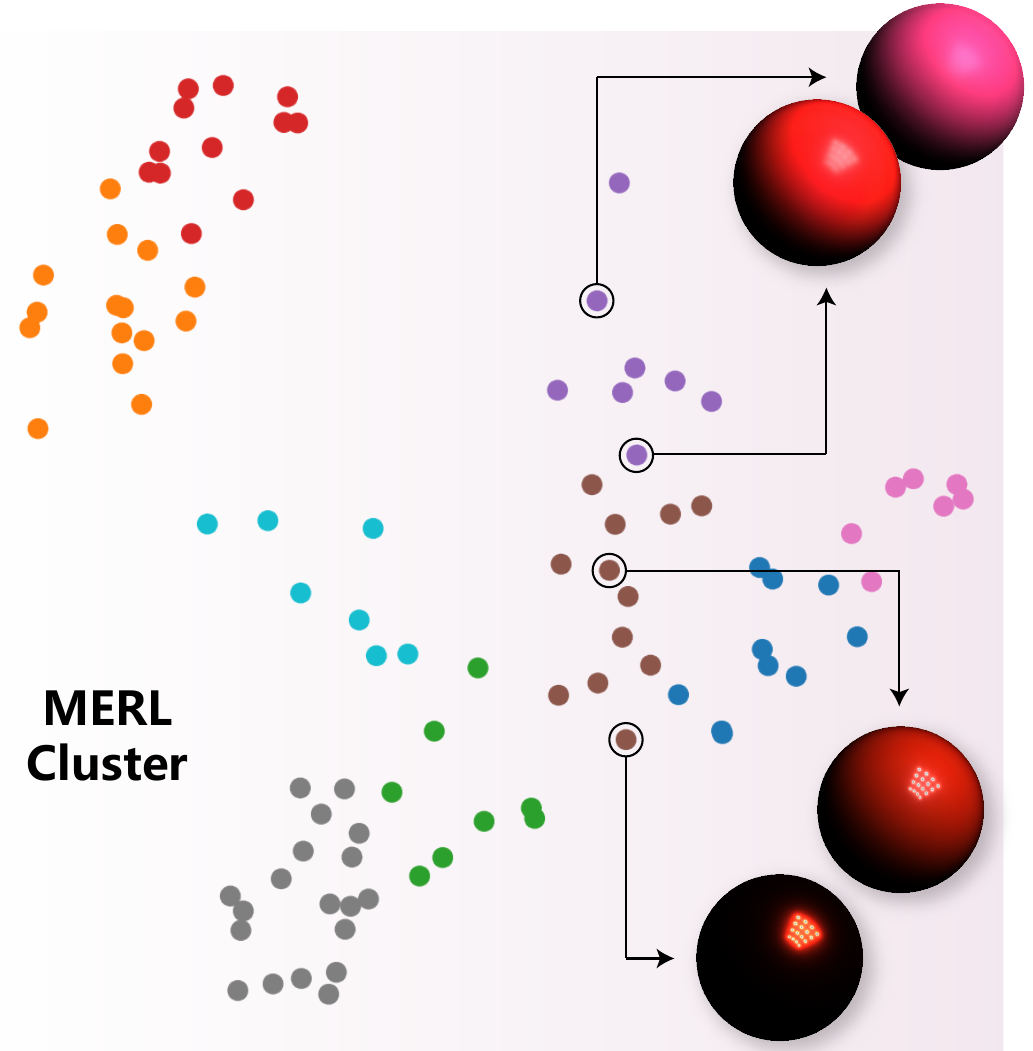}
  \vspace{-7mm}
\end{wrapfigure}
Consider that both the aforementioned codebook and spherical index grid are material-specific. 
Here, we aim at developing a set of \textit{neural reflectance primitives} that 
are learned and shared across multiple materials that have similar characteristics.
Specifically, we first use a convolutional auto-encoder with bottleneck architecture, 
of which the objective is to reconstruct the raw reflectance data. 
The information bottleneck prompts clusters of similar materials.
Thus, a clustering algorithm such as K-Means is then applied to the learned bottleneck codes.
For a cluster with $m$ materials, we instantiate an MLP, a codebook storing the primitives, 
and $m$ spherical index grids $\mathcal{S}_g$, \emph{i.e.} only the low-cost integer indices are exclusive.
Still, both of these modules are trained jointly. 

\begin{figure*}[h]
  \centering
  \includegraphics[width=1.0\textwidth]{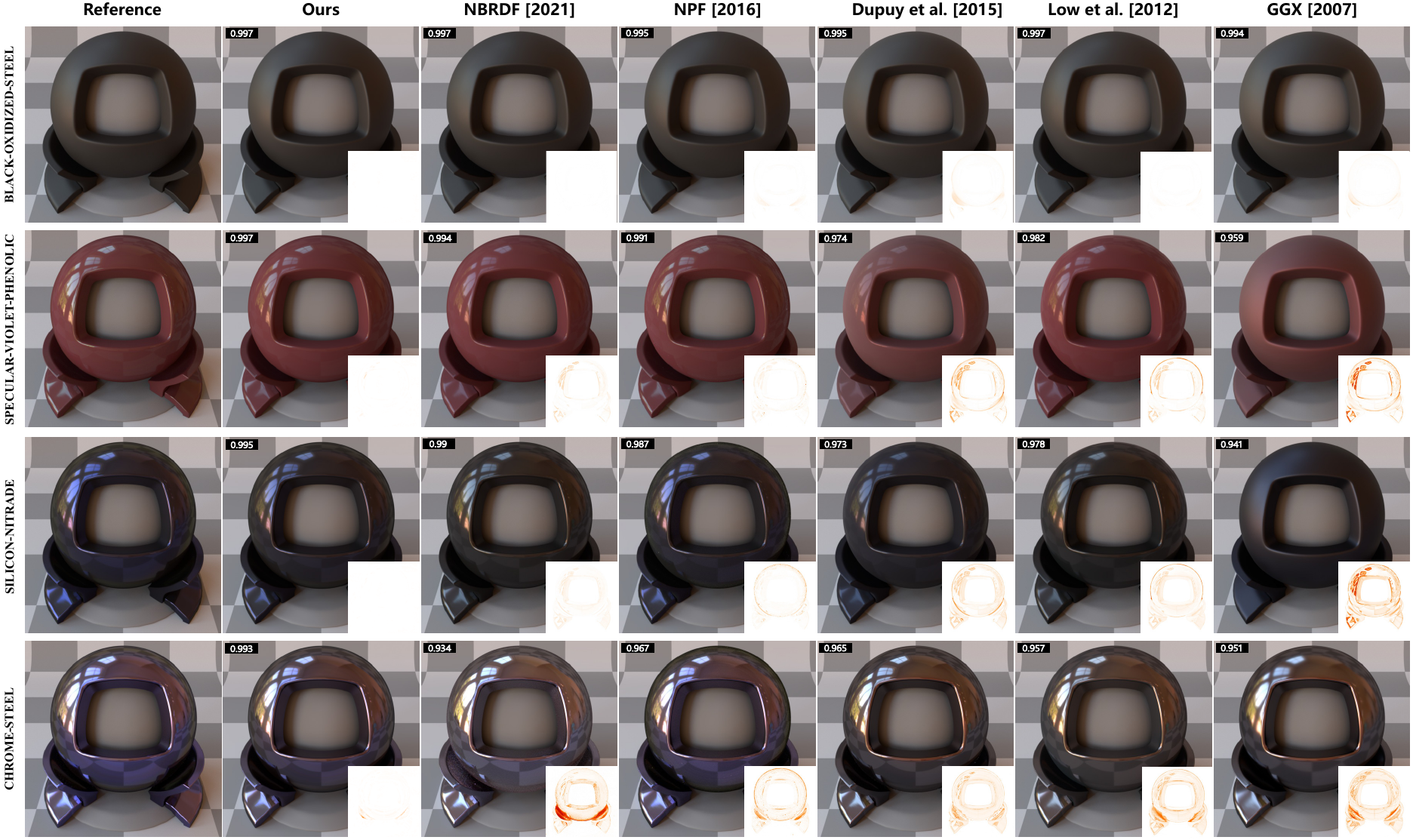}  
  \vspace{-2mm}
  \caption{
    \textbf{Quality comparison in fitting reflectance models to MERL~\shortcite{Matusik:2003} measurements.} 
    The analytical microfacet reflectance models, including GGX and Low et al, 
    struggle in representing the highly frequency changes (SILICON-NITRADE) 
    and often lead to color shift (CHROME-STEEL). 
    Compared with the previous factorization-based methods, 
    NPF~\shortcite{bagher2016non} and Dupuy et al.~\shortcite{dupuy2015extracting}, 
    where the NPF demonstrates fair visual effect, 
    our method shows better qualitative results. 
    We show the SSIM value in top-left corner and SSIM pixel-value in bottom-right, 
    following NBRDF~\shortcite{sztrajman2021neural} for fair comparison. 
  }
  \label{fig_matpreview_isotropic}
  \vspace{-2mm}
\end{figure*}

\section{IMPLEMENTATION}
This section describes the implementation details in terms of 
efficient training, performance at runtime, and extension applications.

\label{sec:implement}
\paragraph{Training}
Given bidirections $\boldsymbol{\omega}_i, \boldsymbol{\omega}_o$ and ground truth reflectance values 
$y$ that are obtained via measurement or simulation,
the optimization problem is:
\begin{equation}
  \underset{\mathcal{D}, \mathcal{S}_g, \Phi}{\arg \min} \ \
  \mathbb{E}_{\boldsymbol{\omega}_{i,o},y}
  || m(\psi(\boldsymbol{\omega}_{i}, \boldsymbol{\omega}_{o}; \mathcal{D}[\mathcal{S}_g]); \Phi) - y||. \label{eq:optim_hard}
\end{equation}
This cannot be solved directly by usual gradient methods since
the indexing $\mathcal{D}[\mathcal{S}_g]$ is non-differentiable.
To this end, we use a proxy soften matrix $\mathcal{C} \in \mathbb{R}^{2n \times 2^b}$ 
from which the integer indices $\mathcal{S}_g = \arg \max_j \mathcal{C}[j]$ can be obtained from 
row-wise argmax, following~\cite{takikawa2022variable}. 
We can then replace the hard indexing $\mathcal{D}[\mathcal{S}_g]$ with a matrix product 
$\sigma(\mathcal{C})\mathcal{D}$, 
where $\sigma(\cdot)$ is the softmax function applied row-wise on matrix $\mathcal{C}$.

For efficient training and better convergence,
we expect the argmax results will almost freeze after a training period and 
the gradient that is backpropagated from a grid point of $\mathcal{S}_g$ 
mainly affects single corresponding primitive.
That is, we should adjust the softness during training.   
Accordingly, we introduce a tunable temperature hyperparameter $\tau$ to the softmax.
We obtain the following optimization problem:
\begin{equation}
  \underset{\mathcal{D}, \mathcal{S}_g, \Phi}{\arg \min} \ \
  \mathbb{E}_{\boldsymbol{\omega}_{i,o},y}
  || m(\psi(\boldsymbol{\omega}_{i}, \boldsymbol{\omega}_{o}; \sigma_\tau(\mathcal{C})\mathcal{D}); \Phi) - y||, \label{eq:optim_soft}
\end{equation}
where the temperature softmax is
$\sigma_\tau (\boldsymbol{x}_i) = \exp(\boldsymbol{x}_i / \tau) / \sum_{j} \exp(\boldsymbol{x}_j / \tau)$.
The hyperparameter $\tau$ controls the softness of the probability distribution.
When $\tau$ gets lower, the biggest value in $\boldsymbol{x}$ gets more probability.
In our case, $\tau$ is initialized with $1.0$ and decreased linearly with the training epoch 
until the minimum $0.5$. 
We also use the straight-through estimator~\cite{DBLP:journals/corr/BengioLC13} to make the loss
be aware of the hard indexing during training, 
\emph{i.e.} we use Equation~\ref{eq:optim_hard} for forward pass and Equation~\ref{eq:optim_soft} 
for backward pass. 
We jointly train the codebook entries, the indices, and the MLP 
by applying Adam~\cite{kingma2014adam}, 
where we initialize learning rate to $10^{-2}$ that is then decayed using multistep scheduler. 


\paragraph{Tiny Neural Network}
A lightweight neural network is made possible when the informative features are fed. 
Our fully connected neural network comprises of four fully connected layers. 
The two hidden layers have 64 neurons each with ReLU activation functions. 

\paragraph{Sphere Grid Pruning}
The measured database is typically acquired using a gonioreflectometer with motorized robotic arms. 
The acquisition setups, such as the angular resolution of sensors and lights, 
vary widely among different databases. 
Due to the uniform pixelation of our spherical grid, 
our framework can adapt to data with different sampling strategies. 
However, there may be some grid points that are never supervised during training. 
For efficiency at runtime and meaningful feature query, we discard these points before deployment. 
That is, a grid point is removed if there is no training direction $\boldsymbol{\omega}$ 
locating at the surrounding four pixels. 
Recall that the HEALPix is hierarchical, \emph{i.e.} a pixel is divided into four at next level, 
we can thus naturally replace these four pixels with their parent, 
which is similar with pruning an octree in~\cite{takikawa2021neural}.  

\begin{figure}[t]
  \centering
  \includegraphics[width=0.99\columnwidth]{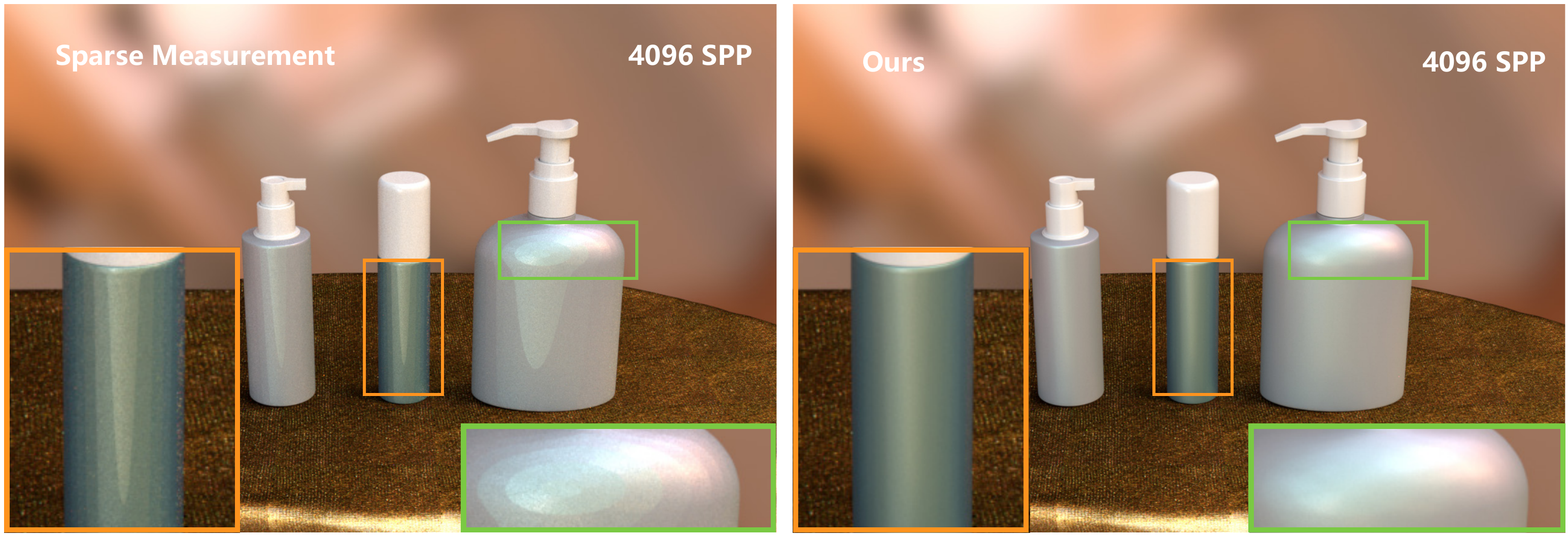}
  \vspace{-2mm}
  \caption{
    \textbf{NeuBRDF trained on sparse measurements.} 
    Our approach can generalize to sparse measurements 
    for both the BRDF (bottles)~\cite{ferrero2014color} 
    and SVBRDF (table cloth)~\cite{merzbach2019learned}. 
    \textit{Bottles:} With our grid pruning, unsupervised pixels are replaced with their parent, 
    resulting in a smooth rendering. 
    \textit{Table Cloth:} NeuBRDF shows faithful rendering effect compared with the measured counterpart.
  }
  \label{fig_sparse_measurement}
  \vspace{-2mm}
\end{figure}

This pruning strategy is useful for isotropic materials or sparse measurements.  
For the isotropic materials, such as those in MERL~\cite{Matusik:2003}, 
only about quarter spherical grids are preserved after pruning. 
\footnote[1]{For isotropic materials, we assign the $\boldsymbol{\omega}_o.\phi$ to azimuth difference with the range $[0, \pi]$
and leave $\boldsymbol{\omega}_i.\phi=0$. Therefore, after pruning, 
about half pixels are kept in south hemisphere and
only pixels surrounding $0$ longitude line are kept in north hemisphere.}
For the sparse measurements, 
our method behaves more like an analytic method, 
where we can simply infer any given direction, rather than 
resorting to time-consuming nearest neighbor search and interpolation like rendering directly with measured data.
Figure~\ref{fig_sparse_measurement} shows a rendering of a NeuBRDF model trained on a 
sparse measurement (448 sampled bidirections)~\cite{ferrero2014color}.



\begin{table}
  \caption{Materials used in our experiments. \textbf{TYPE:} BRDF, BTF and SVBRDF
  are included. 
  \textbf{SAMPLES:} Sparse measurements are involved to assess the generalization ability of NeuBRDF.
  \textbf{SOURCE:} Both the reflectance measured from real-world, simulated by Monte Carlo methods, 
  and evaluated over analytic models are considered.
  (${}^{\bullet}$ isotropic, ${}^{\circ}$ anisotropic.)
  }  
  \vspace{-2mm}
  \label{tab:data_list}
  \begin{minipage}{0.99\columnwidth}
  \begin{center}
  \begin{tabular}{l  l  l  l}
    \toprule
    \textbf{DATABASE}  & \textbf{TYPE} & \textbf{SAMPLES} & \textbf{SOURCE} \\ \hline
    \addlinespace[3pt]
    MERL & $\text{BRDF}^{\bullet}$ & Dense & Measured \\
    RGL  & $\text{BRDF}^{\bullet \circ}$ & Dense & Analytic \\
    LayeredBRDF & $\text{SVBRDF}^{\bullet \circ}$ & Dense & Simulated \\
    Special Coatings & $\text{BRDF}^{\bullet \circ}$ & $448 / 576$ & Measured \\
    UBO2014 & $\text{BTF}^{\circ}$ & $22801$ & Measured \\
    UBOFAB19 & $\text{SVBRDF}^{\circ}$ & $100$ & Measured \\
    \bottomrule
  \end{tabular}
  \end{center}
  \centering
  \vspace{-1mm}
  \end{minipage}
  \vspace{-3mm}
\end{table}

\paragraph{SVBRDF and BTF}
Given 2D location coordinates $\boldsymbol{u} \in \mathbb{R}^{2}$, incident and outgoing directions, 
the spatially varying BRDF (SVBRDF) or bidirectional texture function (BTF)~\cite{dana1999reflectance} 
outputs a reflectance value. 
We discuss these two together since our method focuses on 
evaluation rather than direction sampling.

We propose to use the neural texture~\cite{thies2019deferred} 
as a \textit{plug-in} for our NeuBRDF.
Concretely, we define the neural texture $\mathcal{T} \in \mathbb{R}^{h \times w \times c}$ 
where $c$ is the texture channel.
The neural texture lookup $\varphi(\boldsymbol{u}; \mathcal{T})$ is achieved by grid sampling,
where $\boldsymbol{u}$ is the UV coordinate at the shading point.
The resulting texture vector, together with the feature from spherical grid,
are fed to the MLP $m(\cdot, \Phi)$, which predicts the SVBRDF / BTF value:
\begin{equation}
  f(\boldsymbol{\omega}_i, \boldsymbol{\omega}_o, \boldsymbol{u}) \approx 
  m([\psi(\boldsymbol{\omega}_{i}, \boldsymbol{\omega}_{o}; \mathcal{D}[\mathcal{S}_g]), \varphi(\boldsymbol{u}; \mathcal{T})]; \Phi).
\end{equation}


\begin{figure}[t]
  \centering
  \includegraphics[width=0.99\columnwidth]{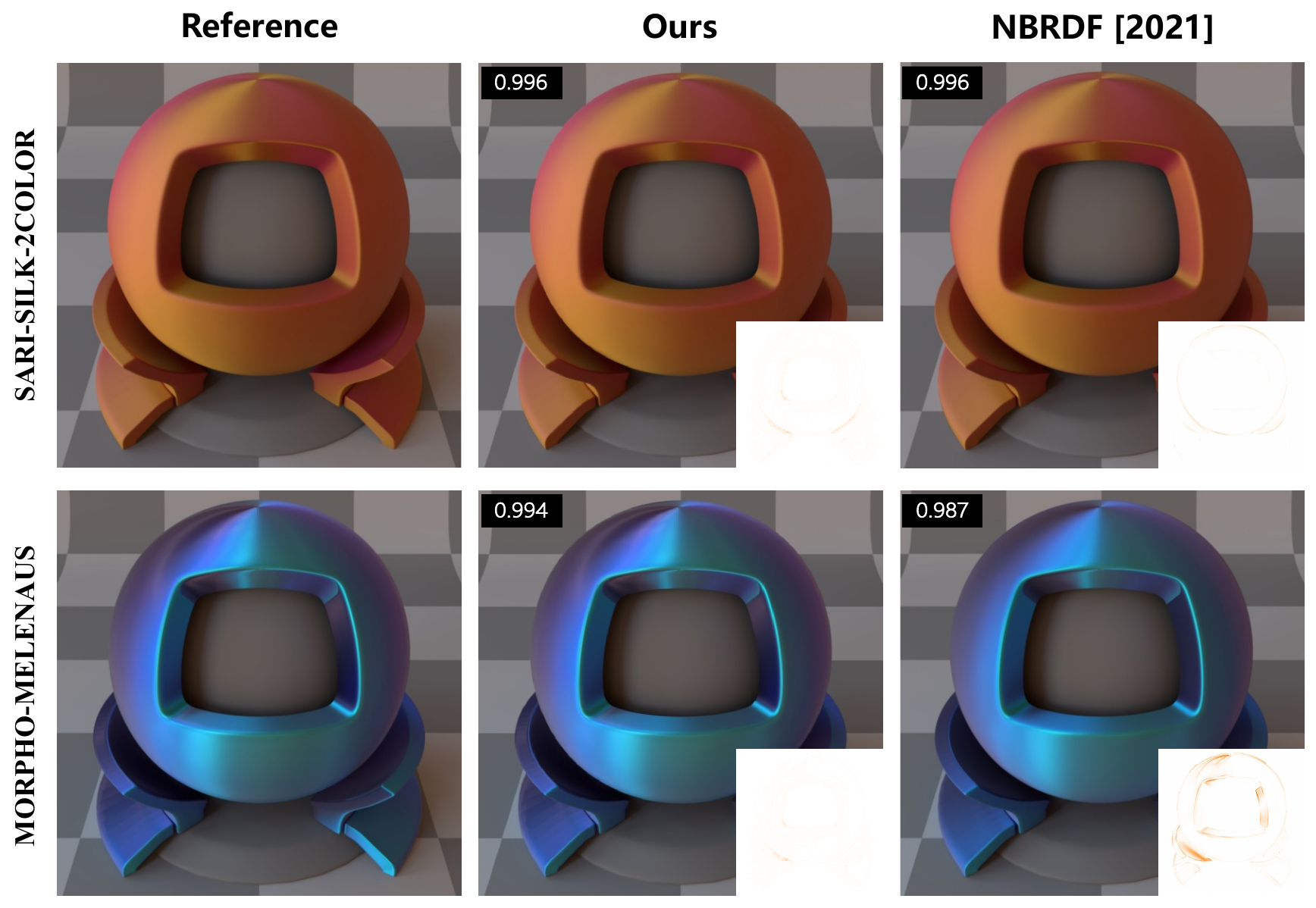}
  \vspace{-2mm}
  \caption{
    \textbf{Quality comparison in representing anisotropic materials from RGL~\shortcite{dupuy2018adaptive} 
    database.} 
    Our method well preserves the highly frequent changes on appearance, 
    especially in the complicated MORPHO-MELENAUS. 
    We show the SSIM value in top-left corner and SSIM pixel-value in bottom-right, 
    following NBRDF~\shortcite{sztrajman2021neural} for fair comparison. 
  }
  \label{fig:iso_matpreview}
  \vspace{-2mm}
\end{figure}

\begin{figure*}[h]
  \centering
  \includegraphics[width=1.0\textwidth]{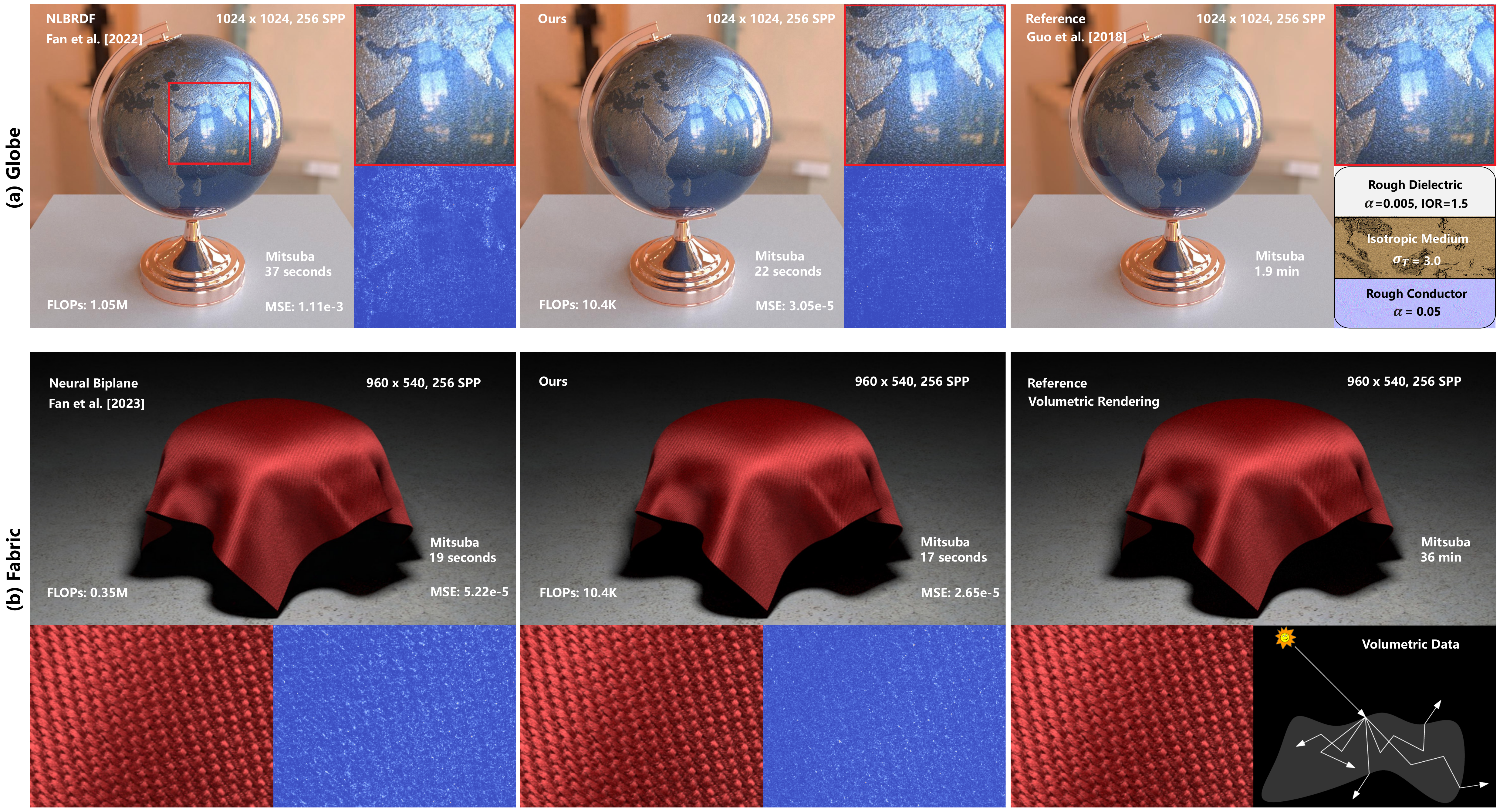}
  \vspace{-2mm}
  \caption{
    \textbf{
    Accelerating physically based simulation.} 
    We also demonstrate the comparison with the recent state-of-the-art neural approach, 
    including NLBRDF~\cite{fan2022neural} and Neural Biplane~\cite{fan2023neural}. 
    \textbf{(a) Simulation from Layered BRDF~\cite{guo2018position}.} 
    Our method requires quarter the time consumption compared to the simulation. 
    \textbf{(b) Simulation from volume rendering.} 
    The fabric is rendered from micro-CT volumetric data, using the microflake phase function. 
    Our method requires almost one percent times compared to the simulation. 
    Additionally, our method demonstrates superior results and performances 
    compared with SOTA neural approaches in both scenes. 
    We implement our methods in Mitsuba (an offline renderer) for fair comparison. 
  }
  \label{fig_globe}
  \vspace{-2mm}
\end{figure*}

\section{Experiments}
\label{sec:experiments}
\subsection{Experimental Setups}
We implement the training in PyTorch~\cite{paszke2019pytorch}. 
After pruning the unused the spherical grid, 
the NeuBRDF is ready for deploying into real-time rendering framework.   
We then deploy the runtime version in Falcor~\cite{Kallweit22} for real-time testing (Figure~\ref{fig:teaser}) 
and in Mitsuba~\cite{jakob2010mitsuba} for fair comparison 
with some methods implemented in this framework. 
We use the following model configuration unless otherwise specified: 
$N_{side}=64, k=16, b=9$ and the MLP has $2$ hidden layers with $64$ neurons.

\paragraph{Datasets.} To demonstrate the versatility of NeuBRDF, we extensively experiment on various materials. 
The databases used in our experiments are listed in Table~\ref{tab:data_list}, 
including isotropic BRDFs, anisotropic BRDFs, and spatially varying BRDFs, 
where the data sources are chosen deliberately to span measured, analytic, and simulated material. 
We also select some very sparse measurement (such as the Special Coating~\cite{ferrero2014color}) 
to demonstrate the generalizability of NeuBRDF and the effectiveness of sphere grid pruning.

\paragraph{Baselines.} We compare our method to several strong BRDF baselines: 
(1) the usual parametrized microfacet model GGX~\cite{walter2007microfacet} and 
the state-of-the-art one proposed by Low et al.~\shortcite{low2012brdf}; 
(2) the recent non-parametric methods relied on factorization~\cite{bagher2016non, dupuy2015extracting}; 
(3) neural approaches for BRDF~\cite{sztrajman2021neural}, 
Layered BRDF~\cite{fan2022neural}, and BTF~\cite{fan2023neural}. 
Furthermore, to demonstrate the application of accelerating physically-based simulation, 
we employ layered material simulator~\cite{guo2018position} and 
volume rendering using microflake phase function 
to generate realistic and complex appearances. 


\begin{table}[h]
  \caption{\textbf{Average image-based losses} of representation methods from 
  Figure~\ref{fig_matpreview_isotropic} over all MERL materials.}
  \vspace{-2mm}
  \label{tab:matpreview_isotropic_error}
  \begin{minipage}{1.0\columnwidth}
  \begin{center}
  \begin{tabular}{l  c  c  c}
    \toprule
    \textbf{Methods} & \textbf{MAE} $\downarrow$ & \textbf{RMSE} $\downarrow$ & \textbf{SSIM} $\downarrow$ \\ 
    \addlinespace[1pt]
    \hline
    \addlinespace[3pt]
    GGX~\cite{walter2007microfacet} & 0.0189 & 0.0206 & 0.969 \\
    Low et al.~\shortcite{low2012brdf} & 0.0080 & 0.0088 & 0.986 \\
    Dupuy et al.~\shortcite{dupuy2015extracting} & 0.0174 & 0.0190 & 0.976 \\
    NPF~\cite{bagher2016non} & 0.0056 & 0.0062 & 0.990 \\
    NBRDF~\cite{sztrajman2021neural} & 0.0028 & 0.0033 & 0.995 \\
    \addlinespace[1pt]
    \hline
    \addlinespace[3pt]
    Ours & \textbf{0.0017} & \textbf{0.0031} & \textbf{0.994} \\ 
    \bottomrule
  \end{tabular}
  \end{center}
  \centering
  \vspace{-1mm}
  \end{minipage}
  \vspace{-3mm}
\end{table}

\subsection{Quality Validation}
It's a standard practice to evaluate BRDF models or tune parameters of analytical models 
on measured reflectance data. 
The fitting results on MERL measurements are demonstrated in Figure~\ref{fig_matpreview_isotropic}. 
Our method achieves the most faithful rendering among both the diffuse, glossy, 
and specular materials. 
The quantitative results on the whole MERL measurement are shown in Table~\ref{tab:matpreview_isotropic_error}. 
We report the mean absolute error (MAE), root mean squared error (RMSE), 
and structural similarity index measure (SSIM) on the rendered image, 
following NBRDF~\cite{sztrajman2021neural}. 

\paragraph{Anisotropic BRDF}
Our sphere data structure is designed to support both isotropic and anisotropic materials. 
This flexibility attribute to the factorized design and grid pruning. 
In figure~\ref{fig:iso_matpreview}, 
we demonstrate the representation quality of anisotropic materials from 
the RGL database. 
NBRDF~\cite{sztrajman2021neural} struggles in representing high-frequency changes 
in some anisotropic BRDFs (such as MORPHO-MELENAUS) 
due to the low frequency spectral bias~\cite{rahaman2019spectral} of the pure-MLP in NBRDF. 
In contrast, benefits from the well-characterized \textit{bidirection encoding}, 
our methods can reproduce the high-frequency reflectance distribution effective.

\paragraph{Sparse Measurement.}
Rendering a sparse measured BRDF directly may encounter visual artifact, 
and a time-consuming nearest neighbor search is necessary for 
rendering smooth results, 
for which the sparse measurements are solely used for rendering but 
are often used for validating analytical reflectance models. 
In order to evaluate the scalability to sparse measurement of NeuBRDF, 
we employ sparse BRDFs database of special coating~\cite{ferrero2014color} 
(hundreds of samples) 
and sparse SVBRDF of UBO2019~\cite{merzbach2019learned} (100 samples per pixel). 
The rendering results are shown in Figure~\ref{fig_sparse_measurement}. 
The smooth surface reflectance benefits from the inherent hierarchy of HEALPix, 
which enables the adaptive grid pruning for arbitrary sparse measurement.

\begin{figure}[t]
  \centering
  \includegraphics[width=0.99\columnwidth]{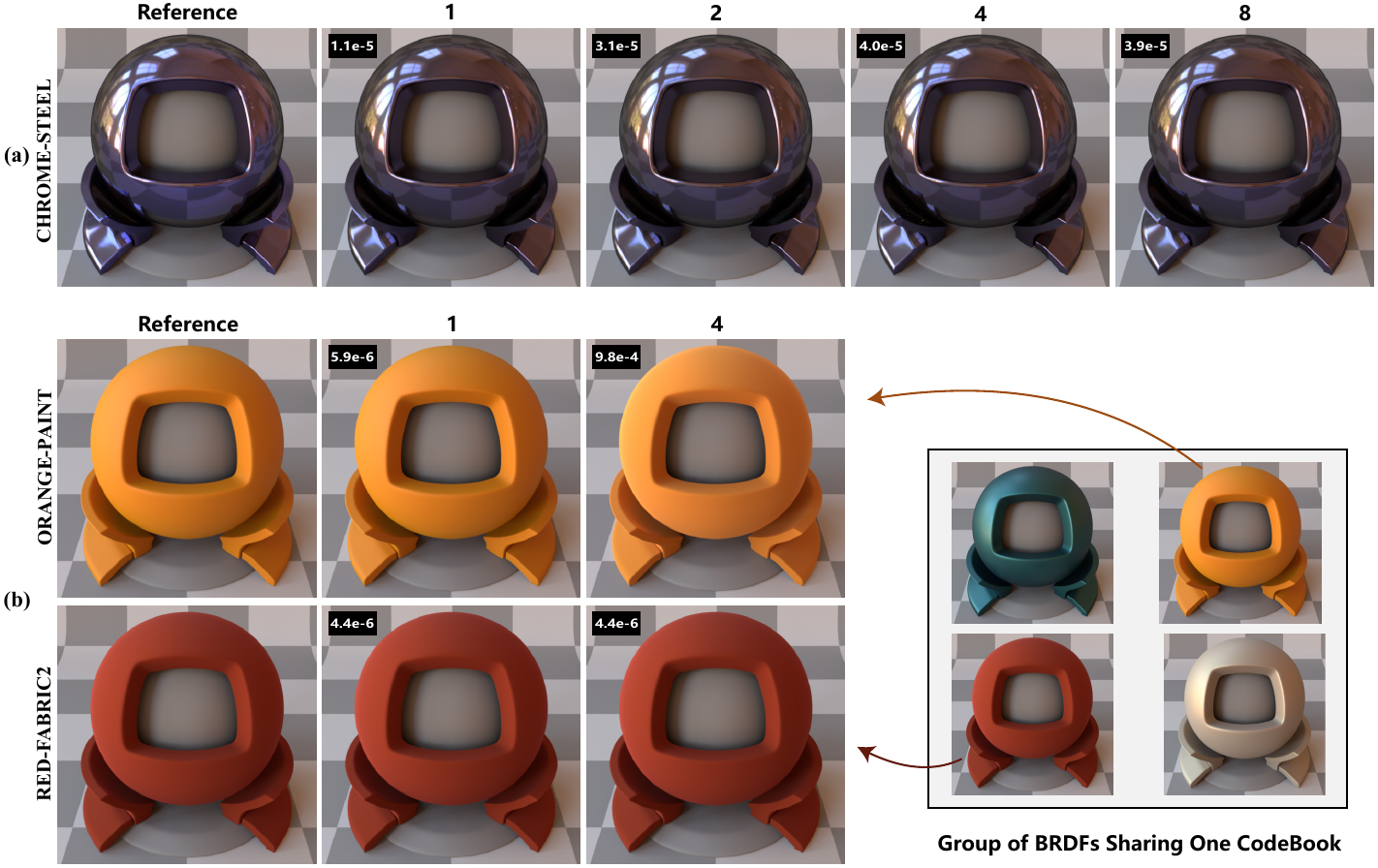}
  \vspace{-2mm}
  \caption{
    \textbf{Sharing codebook across multiple instances.} 
    The digits above images represent the number of instances sharing one codebook. 
    We show the MSE in top-left corner. 
    (a) Representation quality degenerates as the 
    size of belonging material group increases. 
    (b) We demonstrate the ability of neural primitives generalizing to different colors. 
  }
  \label{fig_shared_codebook}
  \vspace{-2mm}
\end{figure}

\begin{figure}[t]
  \centering
  \includegraphics[width=1.0\columnwidth]{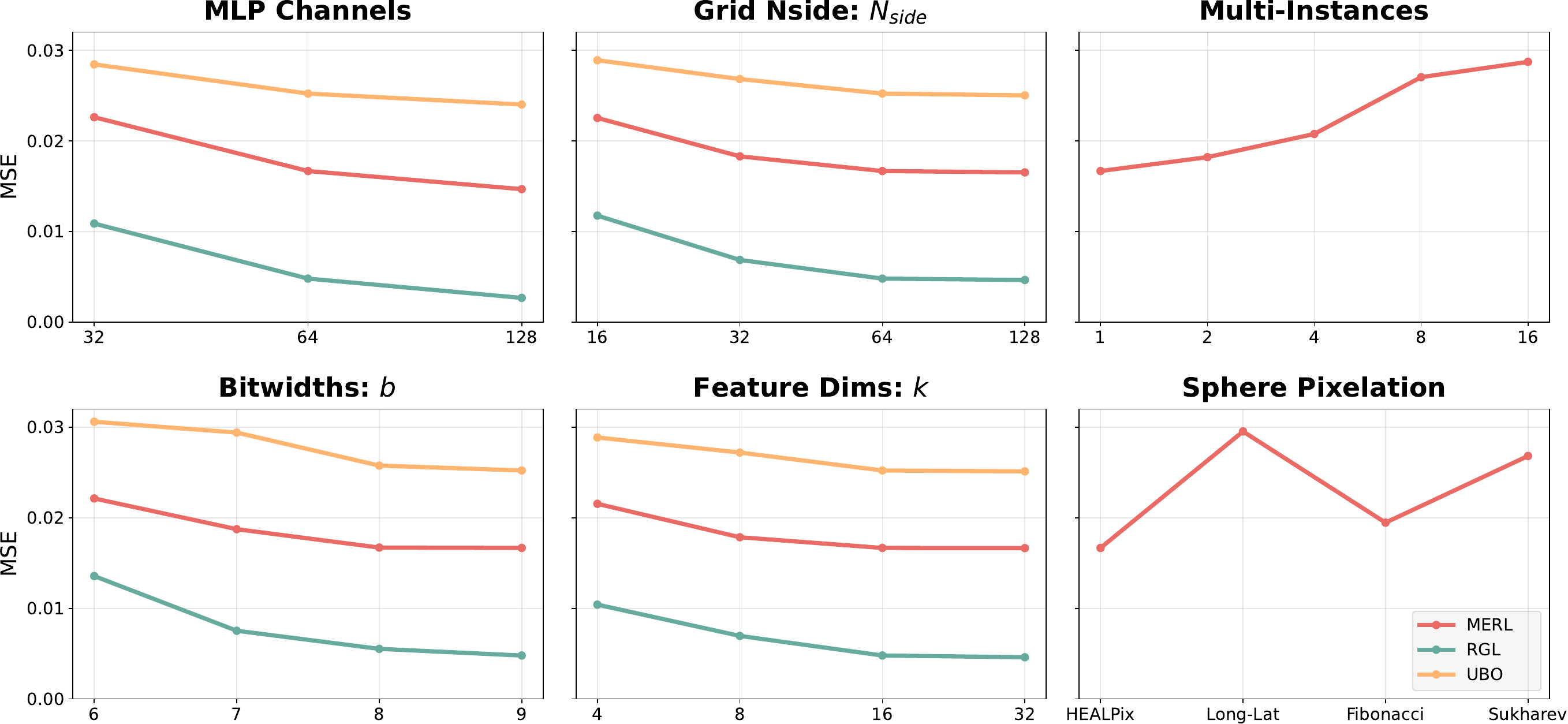}
  \vspace{-2mm}
  \caption{\textbf{Ablation studies on different model configurations.} 
  MSE is computed on the raw data rather than the rendered image. 
  Multiple material instances are selected from a same cluster.
  Sphere pixelations have similar number of grid points for fair comparison.
  }
  \label{fig_ablation_curve}
  \vspace{-2mm}
\end{figure}

\paragraph{Sharing Codebook.}  
The neural reflectance primitives stored in a standalone codebook 
are independent of the per-instance sphere feature-grid, 
which offers the opportunity for maximizing the memory-quality 
by sharing codebook between material instances. 
We illustrate the representation quality in Figure~\ref{fig_shared_codebook}. 
(a) shows the degeneration of rendering quality as 
the number of instances sharing one codebook increase; 
(b) shows the comparison of a cluster of materials represented 
with and without shared codebook. 
Figure~\ref{fig_ablation_curve} (top right) illustrates the overall 
degeneration of MERL when sharing codebook, evaluated on the raw reflectance data.

\subsection{Extension Applications}

\paragraph{Accelerating Physically Based Simulation.}
Reflectance models designed for photorealistic rendering are often 
layered~\cite{Jakob2014Acomprehensive, guo2018position, guillen2020general}. 
By constructing various interface layers and internal media, 
these physical methods can cover a wide range of appearances.
Evaluating such models is usually achieved through light transport between layers, 
simulated by Monte Carlo.
Since our method is agnostic of the internal light-layer interactions 
it can be directly used for accelerating these physical-based methods for various layered configurations. 

In many complex scenes, these layered BRDF methods often use 
albedo texture, normal map, or spatially varying scattering/absorption coefficient in internal layers. 
As a compatible solution, 
we still use the plug-and-play neural texture $\mathcal{T}$. 
The neural texture can be considered as a blend of multiple internal textures. 
With the expense of building the reflectance data (hours) and training ($<1$ hour), 
our method can drastically reduce the cost at rendering; see Figure~\ref{fig_globe} for an illustration. 
Our approach performs surprisingly well in such a complicated scene. 
We use the default base framework together with 
a $800 \times 400 \times 8$ neural texture to capture the spatially varying effects.
These results demonstrate that our approach generalizes well to complex designs 
and can produce highly realistic results in a much smaller computation resource compared 
with either the physical simulations or the recent neural-based approaches.

\begin{figure}[t]
  \centering
  \includegraphics[width=0.99\columnwidth]{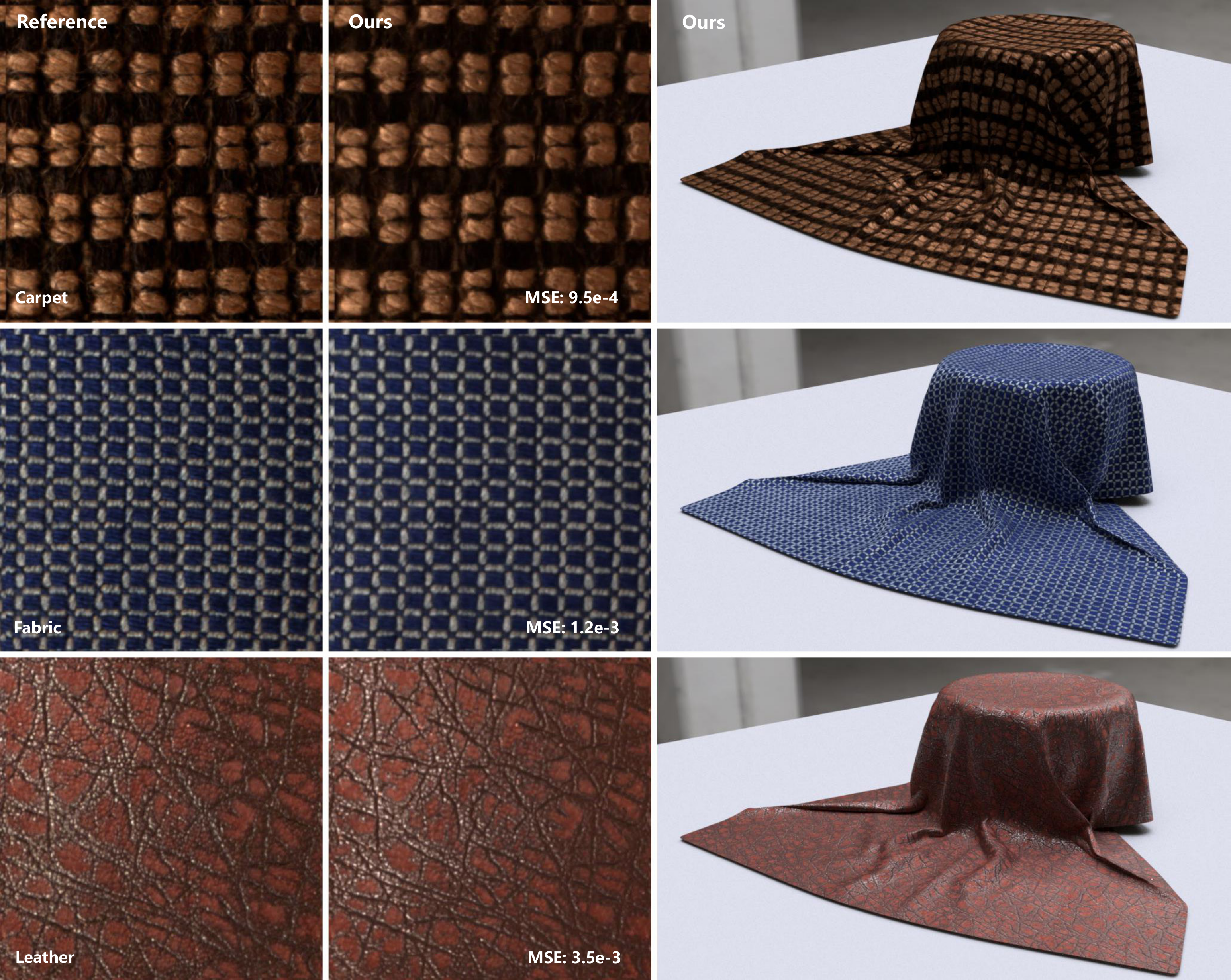}
  \vspace{-2mm}
  \caption{
    \textbf{BTF compression on UBO2014.} 
    Each data contains $400 \times 400 \times 22801$ RGB 
    measurements, which occupancy $\sim500$ MB storage (compressed). 
    Simply with a $128 \times 128 \times 4$ neural texture ($\sim131$KB), 
    NeuBRDF ($\sim95$KB) can capture these spatially varying effects, 
    resulting in a compression ratio of thousands of times. 
    (Storage of NeuBRDF and neural texture are computed in \textit{fp}$16$.)
  }
  \label{fig:svbrdf_ubo}
  \vspace{-2mm}
\end{figure}

\paragraph{BTF Compression}
Tabulating an isotropic BRDF consumes tens of megabytes memory 
and becomes prohibitive for spatially varying material, 
impeding the network transfer and runtime loading. 
Our framework can be easily extended for the compression purpose, 
the BTF compression results are illustrated in Figure~\ref{fig:svbrdf_ubo}. 
The convincing results of NeuBRDF with a naive neural texture plugin suggest 
the feasibility of decoupling spatially varying effect and bidirectional $\omega_{i,o}$ , 
which is also proven in neural biplane~\cite{fan2023neural}. 

Similarly, neural texture compression (NTC)~\cite{vaidyanathan2023random} 
is proposed to compress the texture. 
However, they only operate on the image space and rely on an extra reflectance model to 
consume the decoded image textures. 
Instead, we consider the reflectance model per se. 
We note that a highly specific neural texture like the neural mipmap texture in NTC is also 
compatible with our framework. 

\subsection{Ablation Studies}

\paragraph{Model Configuration}
We investigate the impact of MLP channels, feature-grid resolution ($N_{side}$), 
codebook bitwidth, and codebook feature dimension; 
shown in Figure~\ref{fig_ablation_curve} (left two columns). 
In order to fully validate the model at different configurations, 
we report the MSE on the raw reflectance data rather than the rendered image 
that not yields full coverage of the bidirection space. 

\paragraph{Codebook and Low-bit Quantization}
The primitive indexing is exactly the same with vector quantization 
in the perspective of quantization. 
To this end, we conduct ablation study on (1) the model without codebook, 
where the sphere grid allocates bulky features instead of lightweight integer indices, 
(2) and the model with low-bit quantization. 
The MSE results on MERL raw data and the required parameters 
are shown in Table~\ref{tab:without_codebook}. 
In addition to the memory efficiency, 
the use of codebook can alleviate the overfitting drawback of feature-grid 
and allow the sharing of the primitives across instances as described in previous section. 
We implement low-bit quantization by simulating quantization along the line of 
neural image compression~\cite{balle2016end, vaidyanathan2023random}. 

\begin{table}[h]
  \caption{\textbf{Comparison in the view of quantization.} 
  The codebook is also a quantization technique, 
  for which we compare it with low-bit (4b) quantization. 
  Experimental results show that it consumes less memory and performs better than 
  low-bit quantization in this case. 
  Memory costs are computed in \textit{fp}16 except the low-bit quantization, 
  and the MLP is not taken into account. 
  } 
  \vspace{-2mm}
  \label{tab:without_codebook}
  \begin{minipage}{1.0\columnwidth}
  \begin{center}
  \begin{tabular}{l  c  c  c}
    \toprule
     & \textbf{MSE} $\downarrow$ & \textbf{Memory Cost} $\downarrow$ & \textbf{Ratio} $\downarrow$ \\ 
    \addlinespace[1pt]
    \hline
    \addlinespace[3pt]
    w/o Quantization   & \textbf{0.0160} & 1.60MB & 1 \\
    w Low-bit          & 0.0189          & 0.40MB & 0.25 \\
    w CodeBook (ours)  & 0.0167          & \textbf{0.07MB} & \textbf{0.045} \\
    \bottomrule
  \end{tabular}
  \end{center}
  \centering
  \vspace{-1mm}
  \end{minipage}
  \vspace{-3mm}
\end{table}

%

\paragraph{Sphere Pixelation}
The design of sphere pixelation is critical to the success of our method. 
We demonstrate representing quality in Figure~\ref{fig_ablation_curve} (bottom right), 
where the grid resolutions (or the number of grid points) are managed as close as possible 
for each pixelation. 
Both of the other pixelations are manually tailored for BRDF representation, 
which are similar with that of HEALPix described in Section~\ref{sec:healpix_based_data_structure}. 
Regardless of the exclusive merit of \textit{hierarchy} of HEALPix, 
the model with HEALPix achieves best representation results. 
Although the Fibonacci pixelation is comparable in terms of MSE, 
the indexing time of Fibonacci is about $5ms$ in contrast to the $0.5ms$ of HEALPix. 
We expect the HEALPix with our tailored and delicate adaptation for BRDFs can be 
the basic data structure in material representation, like the Octree in geometry representation.

\paragraph{Runtime Time Breakdown.}
We implement our NeuBRDF in Falcor for real-time testing. 
Please refer to Figure~\ref{fig:teaser} for the time breakdown of our method. 
Evaluating our reflectance models takes about $4.5ms$ in a $5-maxdepth$ path tracer at full 
resolution ($1$ SPP, $1920 \times 1080$), which is agnostic to the material type. 
The bulk of the time is taken by fully connected layers. 
Currently, the fully connected layer is simply implemented in shader without any bells and whistles.

\section{Conclusions}
In this article, we propose a neural reflectance model that is potentially 
a choice for achieving realistic appearance in real-time rendering applications. 
The proposed method is also a unified framework that can represent a variety of materials, 
which can be used for measured data compression, fitting sparse measurements, physical-based methods acceleration  
and so on. 
The key components to the effectiveness consists of the delicately designed sphere 
data structure and neural reflectance primitives. 
In the future, we will exploit more material application based on such architecture, 
\textit{i.e.} spherically distributed primitives with a tiny-MLP decoder, 
such as material editing and material acquisition in inverse rendering. 


\bibliographystyle{ACM-Reference-Format}
\bibliography{sample-base}


\end{document}